\documentclass{article}

\usepackage{comment}

\usepackage{microtype}
\usepackage{graphicx}
\usepackage{subfigure,bbm}
\usepackage{booktabs} 

\usepackage{hyperref}

\def\isaccepted{1} 

\usepackage{icml2025} 

\usepackage{amsmath}
\usepackage{amssymb}
\usepackage{mathtools}
\usepackage{amsthm}
\usepackage[capitalize,noabbrev]{cleveref}

\theoremstyle{plain}

\theoremstyle{definition}

\theoremstyle{remark}

\usepackage[textsize=tiny]{todonotes}

\icmltitlerunning{Empirical Evaluation of Progressive SAEs}

\begin{document}

\twocolumn[
\icmltitle{Empirical Evaluation of Progressive Coding for Sparse Autoencoders}


\icmlsetsymbol{equal}{*}

\begin{icmlauthorlist}
\icmlauthor{Hans Peter}{cs}
\icmlauthor{Anders S{\o}gaard}{cs}
\end{icmlauthorlist}

\icmlaffiliation{cs}{Department of Computer Science, University of Copenhagen, Denmark}

\icmlcorrespondingauthor{Anders S{\o}gaard}{soegaard@di.ku.dk}

\icmlkeywords{sparse autoencoders, dictionary learning, evaluation, interpretability}

\vskip 0.3in
]




\begin{abstract}
Sparse autoencoders (SAEs) \citep{bricken2023monosemanticity,gao2024scalingevaluatingsparseautoencoders} rely on dictionary learning to extract interpretable features from neural networks at scale in an unsupervised manner, with applications to representation engineering and information retrieval. SAEs are, however, computationally expensive \citep{lieberum2024gemmascopeopensparse}, especially when multiple SAEs of different sizes are needed. We show that dictionary importance in vanilla SAEs follows a power law. We compare progressive coding based on subset pruning of SAEs -- to jointly training nested SAEs, or so-called {\em Matryoshka} SAEs \citep{bussmann2024learning,nabeshima2024Matryoshka} -- on a language modeling task. We show Matryoshka SAEs exhibit lower reconstruction loss and recaptured language modeling loss, as well as higher representational similarity. Pruned vanilla SAEs are more interpretable, however. We discuss the origins and implications of this trade-off. 
\end{abstract}

\section{Introduction}

Large Language Models (LLMs) have demonstrated remarkable capabilities across a wide range of tasks \citep{brown2020languagemodelsfewshotlearners,chowdhery2022palmscalinglanguagemodeling,hoffmann2022trainingcomputeoptimallargelanguage,grattafiori2024llama3herdmodels}, but understanding their internal representations remains a significant challenge. Sparse autoencoders (SAEs) \citep{bricken2023monosemanticity,yun2023transformervisualizationdictionarylearning,gao2024scalingevaluatingsparseautoencoders,templeton2024scaling} have enabled extraction of interpretable features from these models at scale, already offering some insights into how LLMs process and represent information. SAEs are computationally expensive to train and run inference on, often prompting developers to train SAEs of varying sizes to balance performance and computational constraints. This is the question we are interested in: How can we efficiently obtain high-fidelity, interpretable SAEs of different sizes for LLMs? 

Our goal is to induce a progressive \citep{skodras2001jpeg2000}, sparse coding that provides us with flexible, dynamic, and more interpretable reconstructions of our representations \citep{gao2024scalingevaluatingsparseautoencoders}. 
In other words, we want to learn a latent space such that for any granularity $G\in \mathbb{N}, G \leq N$, such that the first $G$ dimensions yields good reconstruction performance. We call an SAE with this property a progressive coder, as it allows for graceful degradation of reconstruction quality as we reduce the size of the latent representation and thus the effective number of features used. Throughout this paper, we refer to G as the granularity. By this definition, as the sparse code gets shorter, the computation required for non-sparse matrix multiplication\citep{gao2024scalingevaluatingsparseautoencoders}, is reduced proportionately. For example, if $G = \frac{N}{2}$, the total computation is halved. So is the computation involved in decoding, but this is less important, since encoding is approximately six times as expensive  in the limit of sparsity 
\citep{gao2024scalingevaluatingsparseautoencoders}.

\begin{figure*}[htbp]    
    \centering          
    \includegraphics[width=0.8\textwidth]{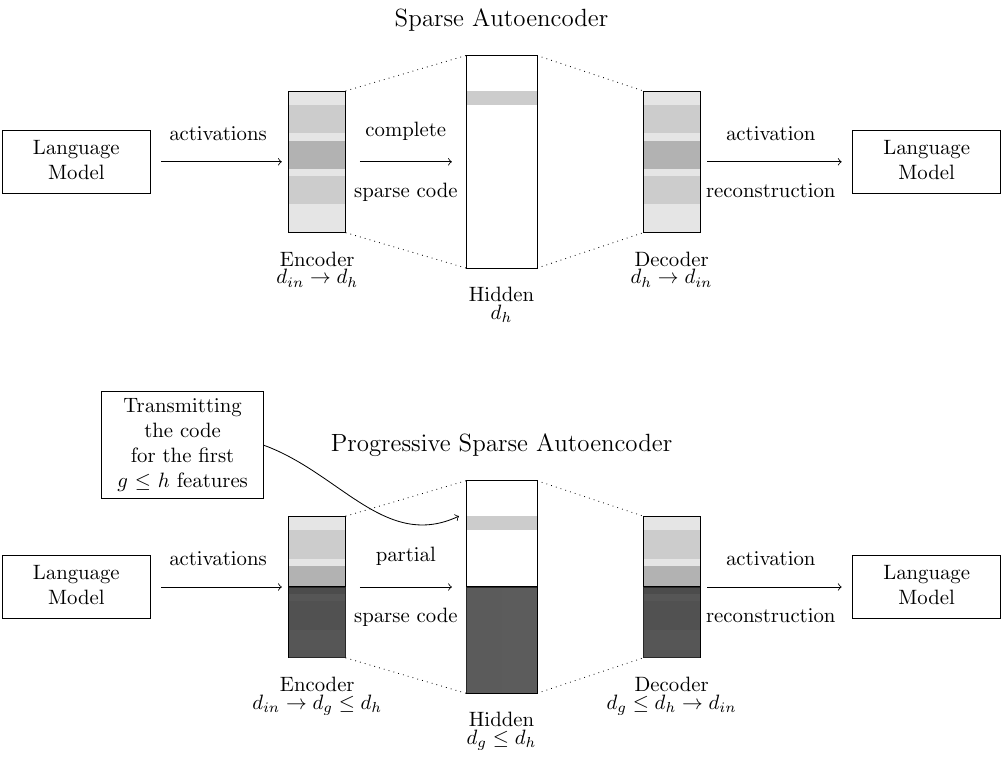}
    \caption{Illustrating progressive coding, the dark part highlight the ressources not used by the model at inference time.}
    \label{fig:progressive-code-vis}    
\end{figure*}

\paragraph{Contributions} We explore two ways of approaching the challenge of inducing progressive SAE coders: (i) Matryoshka SAEs explored independently and concurrently in \citep{bussmann2024learning,nabeshima2024Matryoshka}; 
(ii) pruning vanilla SAEs based on the observed dictionary power law, leveraging their conditional independence. 
Our paper makes the following contributions: (i) We introduce the power law hypothesis for SAE dictionaries. (ii) We introduce a novel baseline method for augmenting pretrained SAEs to become progressive coders. We introduce Matryoshka SAEs, also explored independently and concurrently in \citep{bussmann2024learning,nabeshima2024Matryoshka}. (iii) We compare the two approaches to inducing progressive SAEs across {\em five}~evaluation protocols, including some not previously discussed in the SAE literature.

\section{Background}
\paragraph{SAEs} The superposition hypothesis \citep{elhage2022superposition} posits that neural networks "want to represent more features than they have neurons" \citep{bricken2023monosemanticity}. This phenomenon arises from the fundamental constraint that a vector space can support only as many orthogonal vectors as its dimensionality. To circumvent this limitation, networks learn an overcomplete basis of approximately orthogonal vectors, effectively simulating higher dimensional representations within lower dimensional spaces. 
Such an approximation is theoretically supported by the Johnson-Lindenstrauss Lemma \citep{Johnson1984ExtensionsOL}, which states that for $0 < \epsilon < 1$, any set of $n$ points in $\mathbb{R}^d$ can be embedded into $\mathbb{R}^{O(\epsilon^{-2} \log n)}$ while approximately preserving all pairwise distances between the points up to a factor of $(1+\epsilon)$. In dictionary learning, the goal is to find an overcomplete set of basis vectors $D \in \mathbb{R}^{D \times N}$, with $N >> D$, and a set of representations $R = [r_1, \ldots, r_N]$, where $r_i \in \mathbb{R}^n$, that jointly minimizes reconstruction and sparsity weighted by the sparsity coefficient $\lambda \in \mathbf{R}$:

\begin{equation}
    \arg\min{D, R} \left( \frac{1}{K} \sum_{i=1}^K |x_i - D \cdot r_i|_2^2 + \lambda \mathcal{S}(R) \right)
\end{equation}

where $\mathcal{S}(R)$ is a sparsity measure, commonly implemented as either the L0 pseudo-norm $|r|_0$ or the L1 norm $|r|_1$. However, it remains an open question how to best measure and optimize sparsity \citep{hurley2009comparingmeasuressparsity}. In the interpretability literature, the atoms are most commonly referred to as features, and we use both terms interchangeably. \citet{yun2023transformervisualizationdictionarylearning} were the first to propose dictionary learning for language model interpretability. \citet{bricken2023monosemanticity} and \citet{cunningham2023sparseautoencodershighlyinterpretable} used SAEs to disentangle features in superposition. 
SAEs have weights $W_{\text{dec}} \in \mathbb{R}^{N \times D}$ and $W_{\text{enc}} \in \mathbb{R}^{D \times N}$ and biases $B_{\text{center}} \in \mathbb{R}^{D}$ and $B_{\text{enc}} \in \mathbb{R}^N$. They use an element-wise activation function $\sigma$ such that: 
\begin{align}    
    z &= \sigma((X - B_{\text{center}}) \cdot  W_{\text{Enc}} + B_{\text{Enc}}) \\
    \hat{X} &= (z \cdot W_{\text{Dec}}) + B_{\text{center}}
\end{align}

Different activation functions have been suggested, but the TopK \citep{gao2024scalingevaluatingsparseautoencoders} and JumpReLU \citep{rajamanoharan2024jumpingaheadimprovingreconstruction} activation functions are the most prominent. Our paper exclusively uses the TopK activation function. SAEs are trained by minimizing this loss:
\begin{equation}
    \mathcal{L} = \underbrace{\left(\frac{1}{|D|} \sum_{X \in D} |X - \hat{X}|_2^2\right)}_{\text{reconstruction loss}} + \underbrace{\lambda \mathcal{S}(z)}_{\text{sparsity}}
\end{equation}

\label{mrl}
\paragraph{Matryoshka Representation Learning} (MRL) trains representations in a coarse-to-fine manner, where smaller representations are contained within larger ones. MRL has been applied to NLP \citep{devvrit2024matformernestedtransformerelastic}, in multimodal learning \citep{cai2024Matryoshkamultimodalmodels}, and in diffusion models \citep{gu2024Matryoshkadiffusionmodels}. MRL considers a set $\mathcal{M} \subset \mathbb{N}$ of representation sizes that are jointly learned. Given an input $x$ from domain $\mathcal{X}$, MRL learns a representation vector $z \in \mathbb{R}^{\max(\mathcal{M})}$ such that $z_{1:m_{1}} \subseteq z_{1:m_{2}} \subseteq \dots \subseteq z_{1:m_{n}}$, where each larger representation contains all smaller ones. The representation $z$ is obtained through a neural network $F(\cdot;\theta_F): \mathcal{X} \rightarrow \mathbb{R}^{\max(\mathcal{M})}$ parameterized by $\theta_F$, such that $z := F(x;\theta_F)$. $\mathcal{M}$, typically contains $|\mathcal{M}| \leq \lfloor\log(\max(\mathcal{M}))\rfloor$ elements \citep{kusupati2024Matryoshkarepresentationlearning}. For supervised learning tasks with dataset $\mathcal{D} = \{(x_1,y_1),\ldots,(x_N,y_N)\}$ where $x_i \in \mathcal{X}$ and $y_i \in [L]$, MRL minimizes a linear weighted combination of the loss of the nested models over the dataset $\mathcal{D}$:

{\small
\begin{equation}
    \frac{1}{N} \sum_{i\in[N]} \sum_{m\in\mathcal{M}} c_m \cdot \mathcal{L}(W^{(m)} \cdot F(x_i;\theta_F)_{1:m}, y_i)
\end{equation}}

where $W^{(m)} \in \mathbb{R}^{L\times m}$ represents separate linear models for each nested dimension $m$, and $c_m \geq 0$ denotes the weighted importance of each scale. These weights may be hierarchically structured depending on the task. $F(x_i;\theta_F)_{1:m}$ needs to be computed only once for each $x_i$ by computing $F(x_i;\theta_F)_{1:\max(\mathcal{M})}$, thus this method introduces only the additional overhead of $\sum_{m \in \mathcal{M}} W^{(m)}$ to the forward pass.

\label{dict_power_law}
\section{The Dictionary Power Law Hypothesis}

\citep{li2024geometryconceptssparseautoencoder} found that the eigenvalues of the covariance matrix of the dictionary $W_{dec} \in \mathbb{R}^{N x D}$ follow a power law. This suggests a hierarchical organization of information, where a relatively small number of features capture most of the variance in the data. We examine the mean squared activation value and frequency, as well as replicating their experiment on $10^5$ unseen tokens for three sparse TopK autoencoders of different sizes.

\begin{figure}[H]
    \centering
    \includegraphics[width=3in]{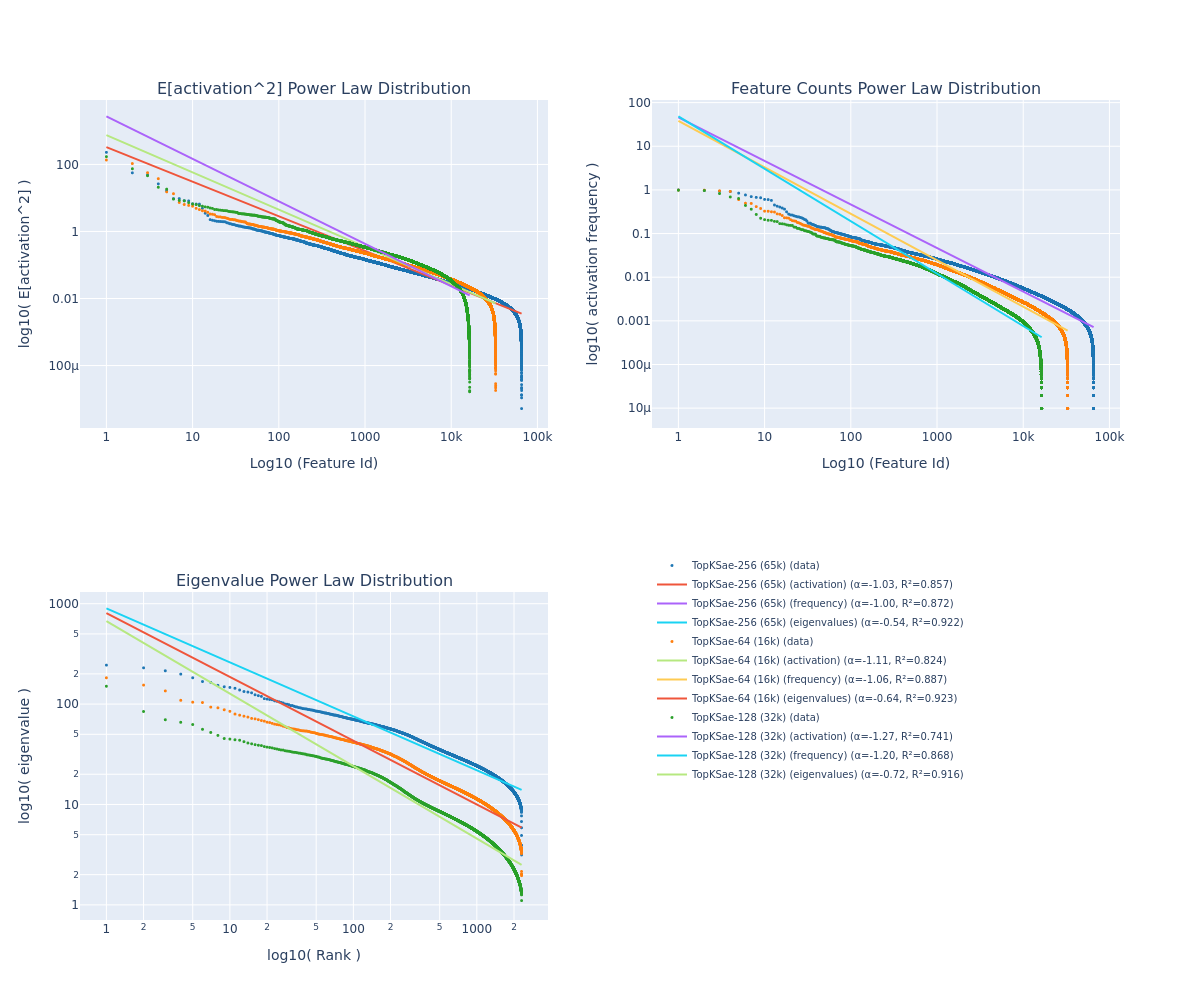}
    \caption{Power law fits for eigenvalues of the covariance matrix, $E[activation^2]$ and activation frequency ($E[\mathbbm{1}{|activation| > 0}]$). We fit a linear regression model to the logarithmically transformed values and display the coefficient and fit for each. We analyze three models of various sizes (65k, 32k, 16k) with consistent sparsity ratios (256-65k, 128-32k, 64-16k).}
    \label{fig:power_laws_combined_plot}
\end{figure}

    The eigenvalues of the decoder matrix's covariance matrix exhibit clear power law decay, with exponents ($\alpha$) ranging from -0.54 to -0.72, and $R^2$ values between 0.916 and 0.922. This indicates a hierarchical structure in the feature space where a small number of directions capture most of the variance. 
    While the squared activation values ($E[activation^2]$) demonstrate approximate power law behavior in their middle range with exponents from -1.03 to -1.27 ($R^2$ values between 0.741 and 0.857), there is a notable deviation in the tail where values decrease more steeply than a power law would predict. 
    Similarly, the frequency of feature activation ($E[\mathbbm{1}{|activation| > 0}]$) follows an approximate power law in its central region with exponents between -1.00 and -1.20 ($R^2$ values between 0.868 and 0.887), but also exhibits a sharp decline in the tail. This indicates that while there exists a hierarchical structure where some features activate much more frequently than others, the least-used features activate even more rarely than a pure power law distribution would suggest. 
Notably, the power law relationships persist across different model sizes, with larger models (TopK-SAE-256) exhibiting slightly less steep decay (smaller absolute $\alpha$ values) compared to smaller models. This consistency across scales and metrics provides strong empirical evidence for the Dictionary Power Law Hypothesis, revealing a robust hierarchical organization of feature importance in SAEs. 

\label{dict_perm}
\section{SAE Dictionary Permutation and Selection}
The dictionary power law suggests that a small subset of features capture most of the important information. We develop a method to identify and prioritize these features by exploiting the permutation invariance of SAEs. 


An important property of SAE features is their conditional independence given the input. Given an input $X \in \mathbb{R}^D $ and an SAE with weights $W_{\text{dec}} \in \mathbb{R}^{N \times D}$ and $W_{\text{enc}} \in \mathbb{R}^{D \times N}$ and biases $B_{\text{center}} \in \mathbb{R}^{D}$ and $B_{\text{enc}} \in \mathbb{R}^{N}$, let $P_{\pi}\in \mathbb{N}^{N \times N} $ be a permutation matrix corresponding to $\pi$. Then, for any permutation $\pi$ of the latent dimensions, the following holds for SAEs: $z = \sigma((X - B_{\text{center}}) \cdot W_{\text{Enc}} + B_{\text{Enc}})$, 
$z' = \sigma((X - B_{\text{center}}) \cdot (W_{\text{Enc}}P_{\pi}) + P_{\pi}B_{\text{Enc}})$, 
$\hat{X} = z W_{\text{Dec}} + B_{\text{center}}$, and $\hat{X}' = z' (P_{\pi}^{-1}W_{\text{Dec}}) + B_{\text{center}}$. This produces identical reconstructions $\hat{X}' = \hat{X}$ for any permutation $\pi$. Each feature activation $z_j$ depends solely on the dot product between the j-th row of $W_{\text{Enc}}$ and the centered input $(X - B_{\text{center}})$, plus its bias term $B_{\text{Enc}_j}$. Independence enables arbitrary reordering of features without affecting overall reconstruction quality.

\begin{figure}[htbp]    
    \centering          
    \includegraphics[width=3in]{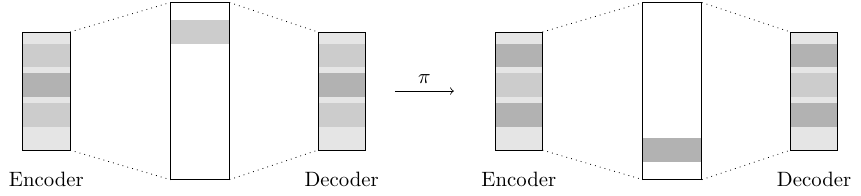}
    \caption{An illustration of dictionary permutation with function $\pi$, Both models will produce the same output given the same input}
    \label{fig:dict-perm-vis}    
\end{figure}

Permutation invariance now enables the conversion of an existing SAE into a progressive coder
by sorting features by descending importance and selecting the first G features at test time. Our objective is to find the permutation $\pi$ that facilitates high-quality reconstruction using only the first $G \in \mathbb{N}, G \leq N$ features of our encoding $z \in \mathbb{R}^N$: $\hat{X}' = z_{1:G} W_{\text{Dec}}[:G, :] + B_{\text{center}}$, where $G$ represents the granularity, or the length of the code the decoder receives. We propose two ranking methods for determining $\pi$: sorting by mean squared activation: $E[activation^2]$; or sorting by mean activation frequency: $E[\mathbbm{1}{|activation| > 0}]$.

\begin{figure}[H]
    \centering
    \includegraphics[width=3in]{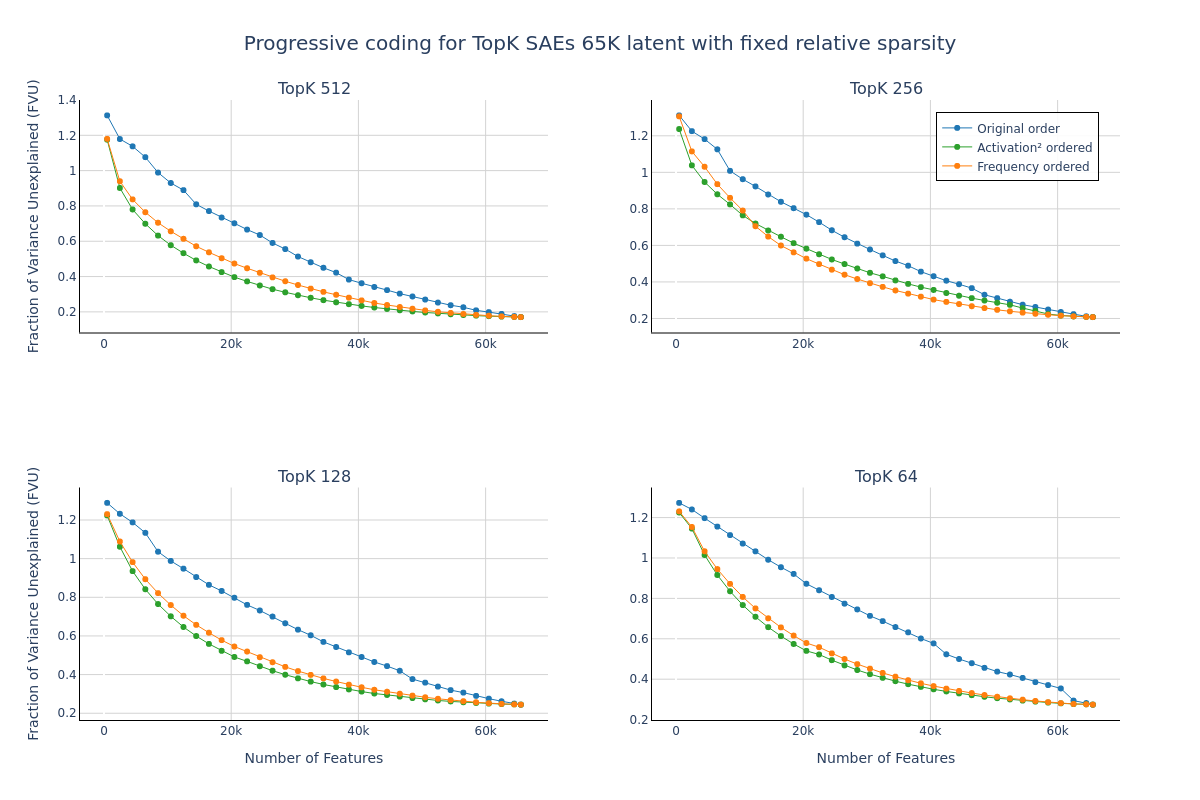}
    \caption{Granularity vs FVU (normalized reconstruction loss) for non-permuted(baseline), permuted based on $E[\text{activation}^2]$ and $E[\mathbbm{1}\{\text{activation} > 0\}]$. Relative sparsity is fixed such that k non-zero / granularity is constant for all granularities}
    \label{fig:topk_power_laws_combined_plot}
\end{figure}

Our results demonstrate that sorting by $E[activation^2]$ consistently achieves the best reconstruction performance across all granularities, and we therefore adopt this ranking method for all subsequent experiments.

\label{Matryoshka_SAE}
\section{Matryoshka SAEs}

We introduce a new method for jointly training nested SAEs by applying principles from MRL \citep{kusupati2024Matryoshkarepresentationlearning}.
Given an SAE with weights $W_{\text{dec}} \in \mathbb{R}^{N \times D}$ and $W_{\text{enc}} \in \mathbb{R}^{D \times N}$ and biases $B_{\text{dec}} \in \mathbb{R}^{N}$ and $B_{\text{enc}} \in \mathbb{R}^D$. Let $M = \{m_1, \ldots, m_k\}$ be the set of representation sizes we want to learn. We denote the forward pass for an SAE for dimension $m_i$, $F_{1:m_i}(\cdot;\theta_F)$ as:

\begin{align}    
    z_{1:m_i} &= \sigma((X - B_{\text{center}}) \cdot W_{\text{Enc}_{:, 1:m_i}} + B_{\text{Enc}_{1:m_i}}) \\
    \hat{X}_{1:m_i} &= z_{1:m_i} \cdot W_{\text{Dec}_{1:m_i, :}} + B_{\text{center}}
\end{align}

\begin{figure}[htbp]    
    \centering          
    \includegraphics{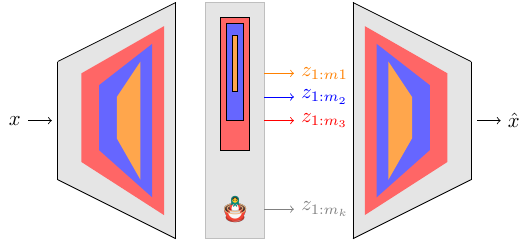}
    \caption{Architectural diagram of the Matryoshka SAE, showing nested latent representations of decreasing dimensionality. The encoder and decoder are shared by each nesting
    }
    \label{fig:Matryoshka-sae}    
\end{figure}

We implement weight sharing in both the encoder and decoder. As $z_{1:m_1} \subseteq z_{1:m_2} \subseteq \ldots \subseteq z_{1:m_k}$ we only have to compute $z_{1:m_k}$ and $z_{1:m_1} \subseteq \ldots \subseteq z_{1:m_k}$ will have been computed. 
This is crucial as the encoding step is the most computationally expensive part of SAE training \citep{gao2024scalingevaluatingsparseautoencoders,mudide2024efficientdictionarylearningswitch}. 
The computational complexity for a naive implementation of an SAE for a batch of size $N$ is dominated by the matrix multiplications $\mathcal{O}(N \cdot D \cdot N)$ for both encoding and decoding, totaling $\mathcal{O}(4 \cdot N \cdot D \cdot N)$ for the forward and backward pass. However, as observed by \citep{gao2024scalingevaluatingsparseautoencoders}, the latent vector is highly sparse, and with an efficient sparse-dense matmul kernel we can compute the decoding step in $\mathcal{O}(N \cdot D  \cdot k)$. By amortizing this cost over 1 encoding step, we can train M nested models for the cost of training the largest one. This gives us a cost of $\approx \frac{\max{M}}{\sum M}$ vs $M$ separate SAEs and as both the encoder and decoder weights are shared, there is no memory overhead. For an efficient implementation of the sparse-dense-matmul kernel, we use the kernel by \citep{gao2024scalingevaluatingsparseautoencoders}. 
Time per step during training increases by $\approx 1.25$ for Matryoshka TopK SAEs, however, this ratio decreases fast with sparsity and larger model sizes. To minimize the amount of dead features, 
we include the auxiliary loss \citep{gao2024scalingevaluatingsparseautoencoders}. We denote the reconstructed activation for granularity $m$ as $\hat{X}_{m}$,
and optimize the following loss:

{\tiny 
\begin{equation}
    \mathcal{L} = \underbrace{\left(\frac{1}{|D|} \sum_{X \in D} \sum_{m \in \mathcal{M}} c_m \cdot |X - \hat{X}_{m}|_2^2\right)}_{\text{reconstruction loss}} + \underbrace{\lambda \mathcal{S}(z)}_{\text{sparsity loss}} + \underbrace{\alpha \cdot \mathcal{L}_{\text{aux}(z)}}_{\text{auxiliary loss}}
\end{equation}
}

where the sparsity and feature activity constraints are only enforced on the full latent representation z, not separately on each nested representation. 



\section{Evaluation and Comparison}

We compare our two approaches, Matryoshka SAEs and column permutation, against baseline TopK SAEs \citep{gao2024scalingevaluatingsparseautoencoders}. We evaluate the performance of our methods by measuring granularity versus reconstruction fidelity at a fixed relative sparsity,\footnote{That is, given our pretrained SAE with dimension $N \in \mathbb{N}, K \in \mathbb{N}, G_1, \dots, G_n \in \mathbb{N} \leq N, K_1, \dots, K_n \in \mathbb{N} \leq K$, we have $\frac{K}{N} = \frac{K_1}{G_1} = \dots = \frac{K_n}{G_n}$
    }, as well as sparsity versus reconstruction fidelity \citep{rajamanoharan2024jumpingaheadimprovingreconstruction}. We train models on 50 million tokens extracted from the second layer residual stream activations (positions 0-512) of Gemma-2-2b \citep{gemmateam2024gemma2improvingopen}, using a random subset of the Pile uncopyrighted dataset.\footnote{Available at https://huggingface.co/datasets/monology/pile-uncopyrighted} The experiments utilized granularities $\mathcal{M} = \{2^{14}, 2^{15}, 2^{16}\}$ and sparsity levels $\{\frac{64}{2^16}, \frac{128}{2^16}, \frac{256}{2^16}, \frac{512}{2^16}\}$, where $k$ is the numerator.

We trained three non-Matryoshka models for each Matryoshka SAE, matching the activation function and dictionary size across granularities $m_i$. Training hyperparameters followed established configurations from prior work\citep{bricken2023monosemanticity,templeton2024scaling,lieberum2024gemmascopeopensparse,rajamanoharan2024jumpingaheadimprovingreconstruction}.

{\tiny
\begin{table}[h]
\tiny
    \centering
    \begin{tabular}{lll}
    \hline
    \textbf{Parameter} & \textbf{Value} & \textbf{Description} \\
    \hline
    Learning Rate & $10^{-4}$ & Optimization step size \\
    Weight Decay & $10^{-2}$ & L2 regularization coefficient \\
    AdamW  & $\beta_1=0.9$, $\beta_2=0.99$, &Momentum and stability terms \\&$\epsilon=10^{-8}$ & \\
    Dictionary Sizes ($\mathcal{M}$) & $\{2^{14}, 2^{15}, 2^{16}\}$ & Nested model granularities \\
    TopK Values ($K$) & $\{64, 128, 256, 512\}$ & Active features per granularity \\
    Auxiliary Loss Scale & $\frac{1}{32}$ & Dead feature regularization \\
    $k_{aux}$ & $\min\{{\frac{d_{sae}}{2}, n\_dead}\}$ & Auxiliary loss feature count \\
    Training Data & 50M tokens & Pile uncopyrighted subset \\
    Context Window & 0-512 & Token positions sampled \\
    \hline
    \end{tabular}
    \caption{Training Configuration for Matryoshka SAEs}
    \label{table:training_hyperparameters}
\end{table}
}

All evaluation metrics were computed on a held-out test set of $10^5$ tokens. As a baseline comparison, we evaluate our results against the JumpReLU SAEs from the GemmaScope family of models \citep{lieberum2024gemmascopeopensparse}. While this comparison provides useful context, several important caveats should be noted:
The training distributions differ, as the exact distribution used in \citep{lieberum2024gemmascopeopensparse} is not publicly documented, and their models are trained on at least an order of magnitude more tokens. Furthermore, the models employ different activation functions (JumpReLU versus TopK), which introduces fundamental architectural differences in how features are encoded and activated.

\paragraph{Progressive coding frontier}
We compute the loss function $L(Z_{1:G})$ across granularities $G \in \mathcal{M} = \{5000, 10000, \ldots\}$, where $G$ represents the dimensionality of the latent space. For each granularity, we maintain a fixed sparsity ratio.  
The evaluation of SAEs remains an open research question \citep{makelov2024principledevaluationssparseautoencoders}\citep{gao2024scalingevaluatingsparseautoencoders}. However, two metrics have emerged as standard in the literature: a) reconstruction loss, measured by FVU(fraction of variance unexplained) or what \citep{gao2024scalingevaluatingsparseautoencoders} calls the normalized mse loss, defined as $\frac{\mathbb{E}[|X - \hat{X}_m|_2^2]}{\mathbb{E}[|X - \bar{X}|_2^2]}$ where $\bar{X}$ is the mean of $X$ over the batch and latent dimension and $\hat{X}_m$ is the reconstruction of $X$ using the SAE with granularity $m$; recaptured LLM loss, i.e., the cross-entropy loss of the unablated model on a dataset divided by the cross-entropy loss when the SAE is spliced into the LMs forward pass: $\frac{\text{Unablated LM loss}}{\text{ablated LM loss}}$.
Importantly, \citep{braun2024identifyingfunctionallyimportantfeatures} demonstrated that discrepancies can arise between these two metrics. As reconstruction loss treats all directions in the activation space as equally important, while in practice some directions may be more functionally significant for the model's downstream performance than others. We find a correlation of about $\approx 0.8$ between the two metrics\ref{fig:metrics_correlation}. We employ representational similarity analysis (RSA) as an additional evaluation metric that bridges between FVU and recaptured LM loss.\footnote{RSA was developed to compare neural representations, but has found applications in machine learning as a measure of second-order isometry 
 \citep{li-etal-2024-vision-language,klabunde2024similarityneuralnetworkmodels}.  
Given $N$ samples of $D$-dimensional activations, RSA forms a matrix $A \in \mathbb{R}^{N \times D}$ containing the original activations. For each representation space, we compute a representational dissimilarity matrix (RDM) using Euclidean distance: $\text{RDM}_A = \left[\sum_{k=1}^D (a_{i,k} - a_{j,k})^2\right]_{i,j=1}^N \in \mathbb{R}^{N \times N}$. The similarity between two representation spaces is the correlation between their RDMs.} For each model $m$, we obtain reconstructed activations $\hat{A}_m \in \mathbb{R}^{N \times D}$ by passing the original activations $A$ through the model. We then compute RDMs for both the original and reconstructed activations. The RSA score for model $m$ is computed as the Pearson correlation between the upper triangular elements of the original and reconstructed RDMs: $\text{RSA}_m = \text{corr}(\text{triu}(\text{RDM}_A), \text{triu}(\text{RDM}_{\hat{A}_m}))$. We examine how these metrics correlate in Appendix \ref{fig:metrics_correlation}.

\begin{figure}[H]
        \centering
        \includegraphics[width=3in]{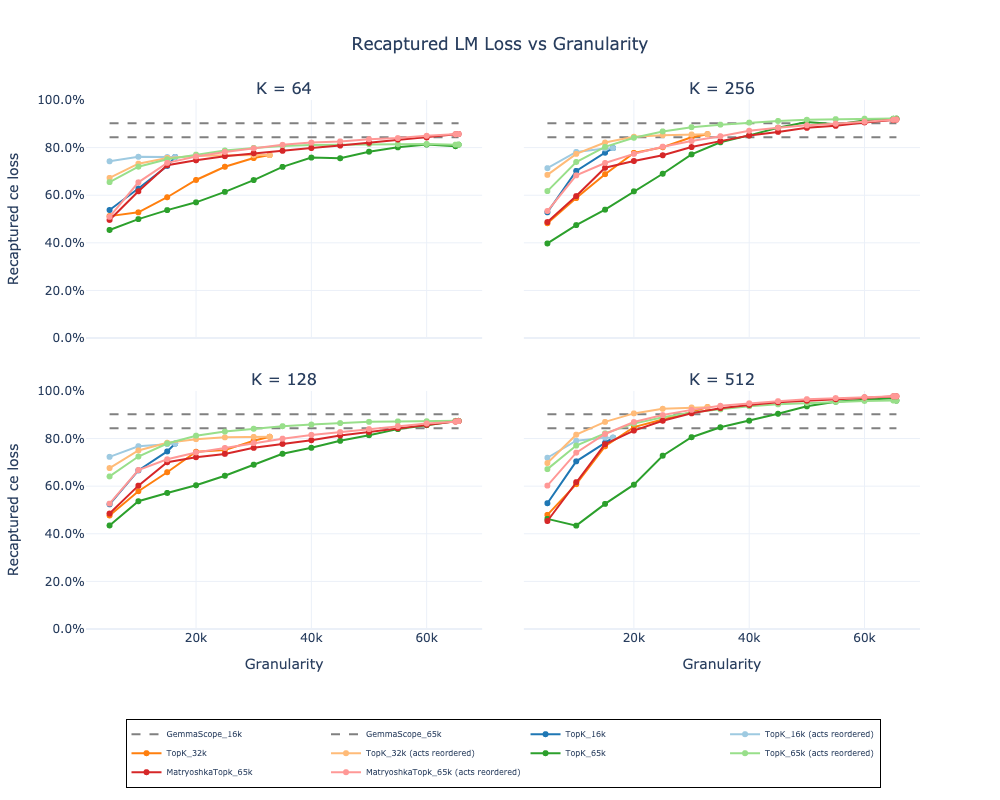}
        \caption{Mean cross-entropy loss per token for gemma-2-2b divided by the cross-entropy loss using the SAE reconstruction, computed over $10^5$ tokens on the pile-uncopyrighted dataset. K refers to the sparsity mechanism, for the topk activation function, which all our models use, in topk all but the K largest features are used, all other are set to zero}
        \label{fig:topk_progressive_coding_ce_granularity_a}
    \end{figure}
    
\paragraph{Results} For all granularities, the Matryoshka SAE outperforms the baseline SAEs as well as the baseline column permuted SAE on the granularity-versus-reconstruction fidelity frontier. This suggests that the Matryoshka SAE has learned to be a more efficient progressive coder. We also observe that applying column permutation approach to Matryoshka SAE increases performance further, although we believe this impact is greatly diminished when using more granularities.

\begin{figure}[H]
        \centering
        \includegraphics[width=3in]{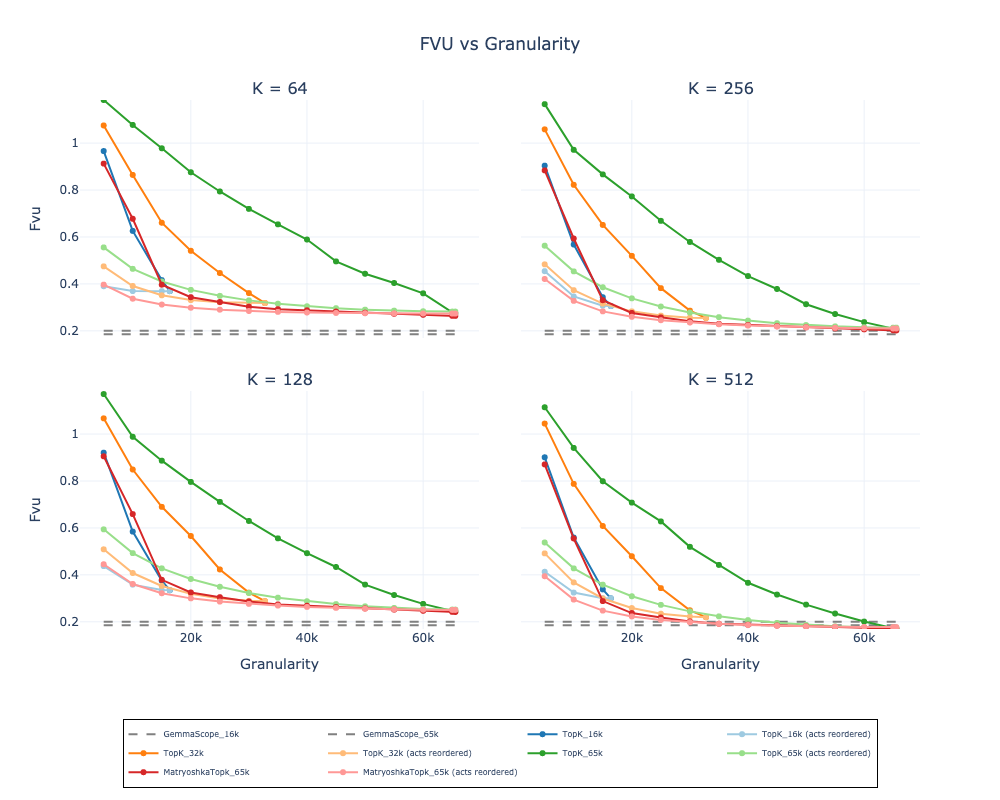}
        \caption{Fvu per token for gemma-2-2b divided by the cross-entropy loss using the SAE reconstruction, computed over $10^5$ tokens on the pile-uncopyrighted dataset.}
        \label{fig:topk_progressive_coding_ce_granularity_b}
    \label{fig:topk_progressive_coding_combined}
\end{figure}

\begin{figure}[H]
    \centering
    \includegraphics[width=3in]{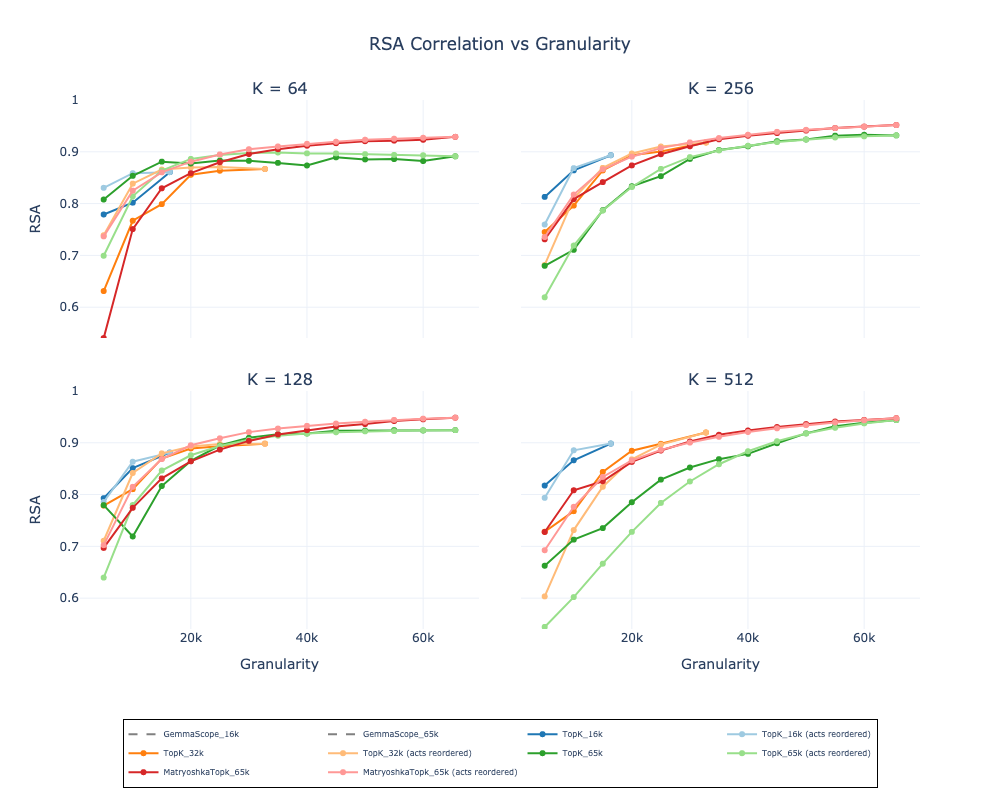}
    \caption{RSA per token for gemma-2-2b divided by the cross-entropy loss using the SAE reconstruction, computed over $10^5$ tokens on the pile-uncopyrighted dataset.}
    \label{fig:rsa_progressive_coding_ce_granularity}
\end{figure}

\paragraph{Sparsity-Fidelity Frontier}
\label{sec:sparsity_vs_fidelity_frontier}
Next, we evaluate the sparsity vs fidelity frontier for our different approaches. For a fixed dictionary size, we evaluate models with four different sparsity levels using the hyperparameters described in Table \ref{table:training_hyperparameters}. We measure sparsity using $\mathbb{E}[|z|_{0}]$, which equals $k$ when using the TopK activation function that fixes the number of non-zero latents. We evaluate models using the same performance metrics as in Section \ref{eval_progressive_coding}, testing each model at full capacity (i.e., using all available features with granularity $G$ equal to the model's total dimension $N$).

\paragraph{Results}

\begin{figure}[H]
    \centering
    \includegraphics[width=3in]{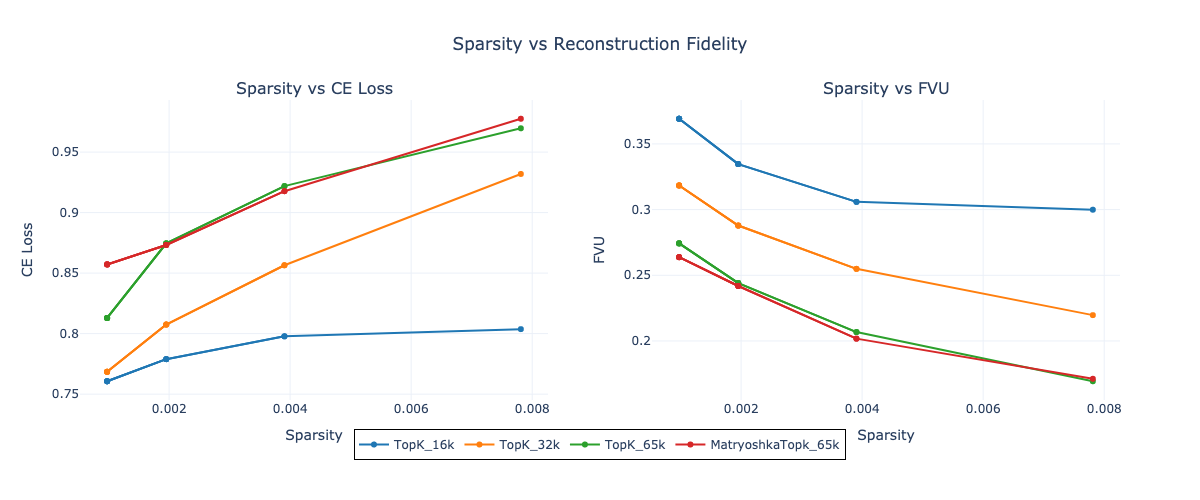}
    \caption{Sparsity vs Reconstruction fidelity (FVU)}
    \label{fig:topk_sparsity_frontier_fvu}
    \caption{Sparsity frontiers for different metrics computed over $10^5$ tokens from the pile-uncopyrighted dataset.}
    \label{fig:sparsity_frontiers}
\end{figure}

We find that Matryoshka SAEs closely track the performance of a baseline autoencoder of the same size both in terms of recaptured downstream cross-entropy loss and reconstruction loss \ref{fig:topk_sparsity_frontier_fvu}.

Next we compare the performance of MatryoshkaSAEs and TopKSAEs using only the first 16K and 32K latents($G$). We find that applying either of our two methods (Matryoshka SAE or column permutation) to a larger SAE, and using the first $n$ latents when reordering is applied, achieves performance comparable to training an SAE of that same size from scratch. This suggests that given a fixed computational budget, it may be more efficient to train one large SAE and subsequently distill it into smaller ones, rather than training multiple SAEs with less compute.

However, this effect becomes less pronounced as the ratio of granularity to model size decreases. While both the 65K Matryoshka SAE and TopK permuted SAE outperform a baseline 16K TopK SAE when using only their first 16K latents, they are in turn outperformed by the 32K SAE with reordering at the 16K or 10K granularity level.

This is likely attributable to the phenomenon of feature-splitting \citep{bricken2023monosemanticity}, where a single latent in a smaller SAE is split into multiple latents in a larger one. Thus, although we observe features follow a power law, as our latent space grows, the importance of any given feature may be gradually diluted as it becomes distributed across multiple features. In Section \ref{sec:limitations_and_future_work}, we propose future approaches that might recover the performance lost from feature splitting.

\begin{figure}[H]
    \centering
    \includegraphics[width=3in]{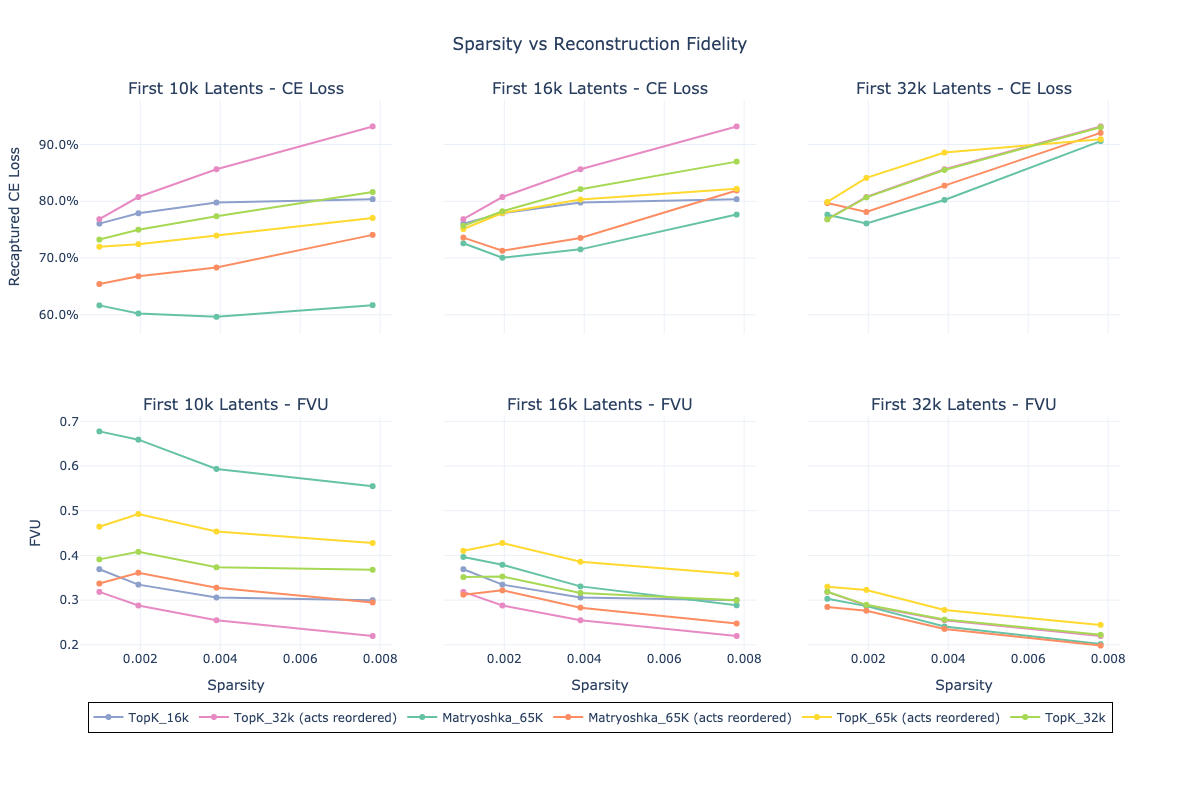}
    \caption{Sparsity vs Reconstruction fidelity for models, only using the first 10k, 16k or 32k latents. Lower fvu is better, higher recaptured ce-loss is better.}
    \label{fig:topk_sparsity_frontier_fvu_16k_32k}
\end{figure}

\label{sec:interp}
\paragraph{Interpretability} As Matryoshka SAEs are a new method for training SAEs, we find it important to evaluate whether this architecture compromises on interpretability. Our other approach, column permutation, is exempt from this analysis, as this method does not change features themselves only their ordering. We evaluate the interpretability of our architecture using two methods from the
automated interpretability library 'sae-auto-interp'\citep{eleutherai2024saeautointerp}: simulation scoring and fuzzing. We evaluate the interpretability of our models by measuring how well a large language model can predict the activation value of our features, given an LM-explanation generated from a training set of examples. This method was first proposed by \citep{bills2023language}, and it measures how correlated an LLM's guess of an activation is with the ground truth activation. We group our activations into 10 quantiles of 50 features based on their firing frequency after having filtered out dead features\footnote{features with a firing frequency of 0}. We compute the Pearson correlation between the activations of the SAE feature in question and the LM simulated activation. We use sequences of context length 32, 10 test samples and 20 training samples used to generate the LM-explanations. All experiments are performed using Llama-3-1-70B \citep{grattafiori2024llama3herdmodels}. We compare the results for our Matryoshka SAE against the baseline Topk SAE, as well as GemmaScope \citep{lieberum2024gemmascopeopensparse} JumpReLU SAEs with approximately the same dictionary size and sparsity level. We compare these against a randomly initialized SAE.

\begin{figure}[H]
    \centering
    \includegraphics[width=3in]{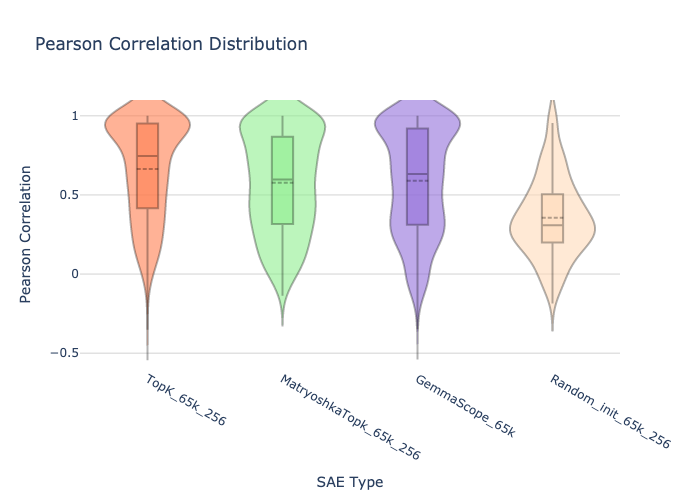}
    \caption{The Pearson correlation between Llama-3 simulated and ground truth activations. The dashed lines represent the mean per SAE type. 
    Values above 1 are an artifact of the kernel density estimation process}
    \label{fig:simulator_interpretability}
\end{figure}

\paragraph{Results} We find that although the Mean Pearson Correlation is meaningfully higher than the randomly initialized SAE\ref{fig:simulator_interpretability} and {\em on par} with the GemmaScope models, our Matryoshka SAE underperforms the baseline TopK SAE models. 

To get a better grasp of exactly which features become less interpretable, we visualize the distribution of Pearson correlations for different granularities. 

\begin{figure}[H]
    \centering
    \includegraphics[width=3in]{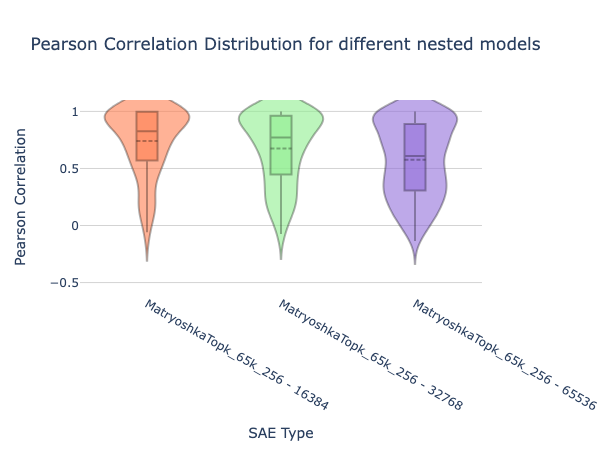}
    \caption{The Pearson correlation between Llama-3 simulated and ground truth activations for different granularities of a Matryoshka SAE. Note that the granularities of 16k are a subset of 32k etc. 
    Values above 1 are an artifact of the kernel density estimation process}
    \label{fig:simulator_interpretability_Matryoshka}
\end{figure}

We find that the innermost granularities are meaningfully more interpretable than the outermost, going from a mean correlation of 0.57 to 0.74. We posit that this occurs as the model, through the Matryoshka loss function \ref{Matryoshka_SAE}, becomes incentivized to effectively put the most meaningful features in the first part of the $W_{dec}$ matrix.\\[1pt]

\noindent Next we evaluate our models using fuzzing, a token-level evaluation technique introduced by \citep{paulo2024automaticallyinterpretingmillionsfeatures}. In fuzzing, potentially activating tokens are highlighted within example sentences, and a language model is prompted to identify which markings are correct. Unlike simulation scoring \citep{bills2023language}, which requires predicting continuous activation values, fuzzing frames the problem as a binary classification task\citep{paulo2024automaticallyinterpretingmillionsfeatures}: Determining whether a token triggers a given feature or not. 

\begin{figure}[H]
    \centering
    \includegraphics[width=3in]{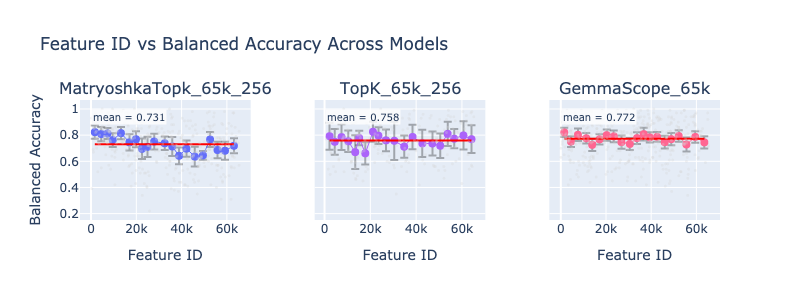}
    \caption{Balanced accuracy for feature indices grouped into quantiles 0-100 for 400 randomly selected features}
    \label{fig:fuzzing_interpretability_quantile_balanced_accuracy}
\end{figure}

\paragraph{Results} We plot the mean balanced accuracy of feature quantiles by frequency in Figure~\ref{fig:fuzzing_interpretability_quantile_balanced_accuracy}. 

Matryoshka SAEs slightly underperform on this task: The first latents seem to perform better than the average, but scores quickly drop.

\section{Discussion: Scaling and Granularities}

An obvious question is, given a large SAE, how well can the performance of the model be predicted when only the G first elements are considered? 
Specifically what is the interaction between model size (N), granularity (G), and sparsity (K) as we scale? We develop empirical scaling laws following the methodology established by \citep{kaplan2020scalinglawsneurallanguage} by modelling how reconstruction loss(FVU) scales with model size and sparsity for baseline TopKAutoencoders with dictionary permutation/reordering applied. Building on the work of \citep{gao2024scalingevaluatingsparseautoencoders}, 
we extend their formulation with two terms: $\beta_g \log(g)$ for the direct effect of granularity, and $\gamma_g \log(k)\log(g)$ for its interaction with sparsity.

\begin{equation}\footnotesize
\begin{split}
    L_{\text{progressive}}(n, k, g) = & \exp(\alpha + \beta_k \log(k) \\ &  + \beta_n \log(n) + \beta_g \log(g) \\ & + \gamma_n \log(k)\log(n) \\ & \underbrace{+ \gamma_g \log(k)\log(g))}_{\text{loss}} \\
    & \underbrace{+ \exp(\zeta + \eta \log(k))}_{\text{irreducible loss}}
\end{split}
\end{equation}

We fit our scaling law using validation data from 16k, 32k, and 65k TopK SAEs with sparsity levels described in \ref{table:training_hyperparameters}, inducing parameters: $\alpha = -3.60$, $\beta_k = 0.69$, $\beta_n = 0.19$, $\beta_g = 0.08$, $\gamma_n = 0.02$, $\gamma_g = -0.10$, $\zeta = -2.13$, $\eta = -0.13$ with $R^2 = 0.978$ in log-log space.

\begin{figure}[H]
    \centering
    \includegraphics[width=3in]{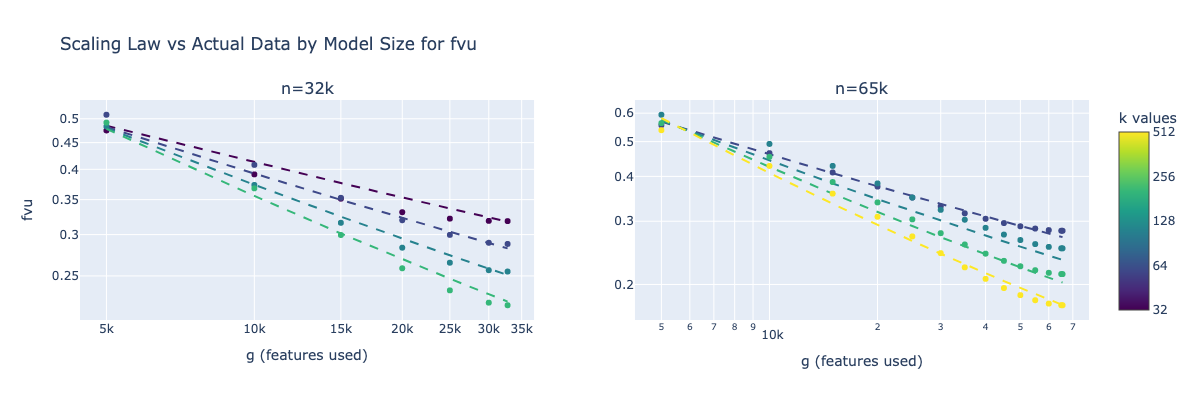}
    \caption{Loss vs.~predicted loss for SAE (32k and 65k latents)}
    \label{fig:scaling_laws}
\end{figure}



Substantial evidence supports that Matryoshka SAEs learn a hierarchy of features, placing the most important features in the first $m_k$ columns of the decoder. 
Earlier work on MRL\citep{devvrit2024matformernestedtransformerelastic} has suggested sampling granularities during training. This idea is very similar to nested dropout \citep{rippel2014learningorderedrepresentationsnested}, where higher-dimensional components of the representation are stochastically dropped out to encourage ordering of dimensions by importance. We apply this approach to Matryoshka SAEs. We hypothesize that sampling granularities dynamically would further improve progressive coding abilities, by learning a finer-grained hierarchy of features. We sample $m_i \sim \mathcal{U}(1, N)$ uniformly at each training step, where $N$ is the maximum dimension of our latent space. 

\begin{figure}[H]
    \centering
    \includegraphics[width=3in]{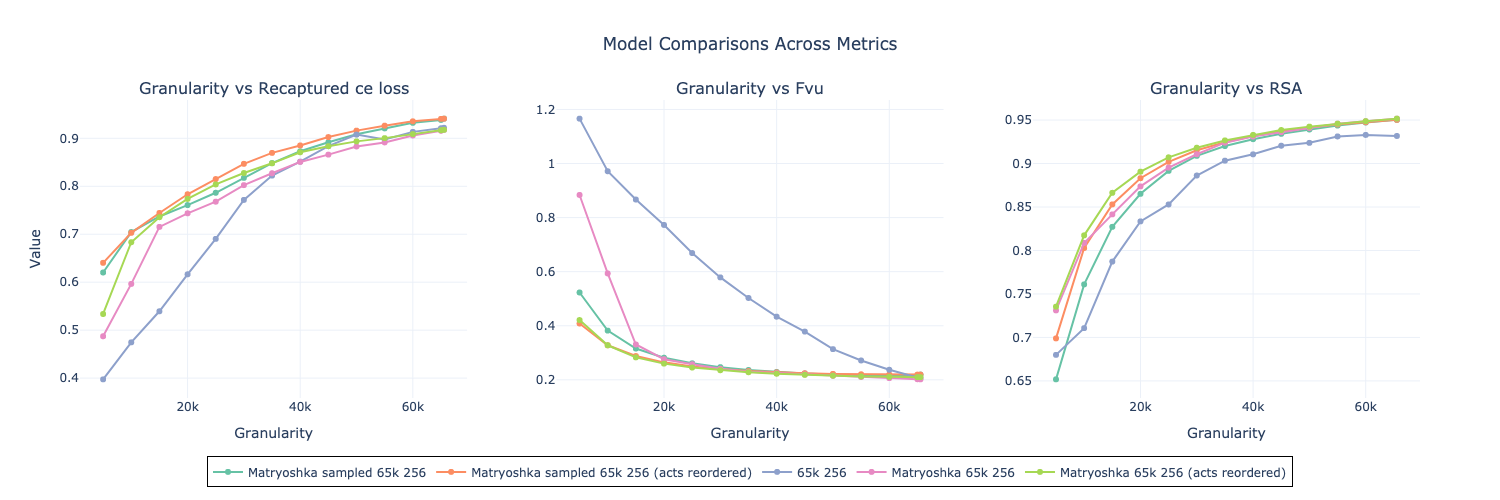}
    \caption{Sampled, non-sampled Matryoshka and baseline ($10^5$t)}
    \label{fig:sampled_progressive_code_ce_loss}
\end{figure}

Sampling improves Matryoshka SAE metrics. 
The sampled Matryoshka SAE concentrates most of its activation mass in the first features, while the non-sampled exhibits distinct plateaus for each granularity level. The baseline TopK SAE shows a more uniform distribution of activation mass across its feature space. This suggests that both Matryoshka variants learn to concentrate important features early in their latent space, but the fixed granularity version creates more structured groupings.

\begin{figure}[H]
    \centering
    \includegraphics[width=3in]{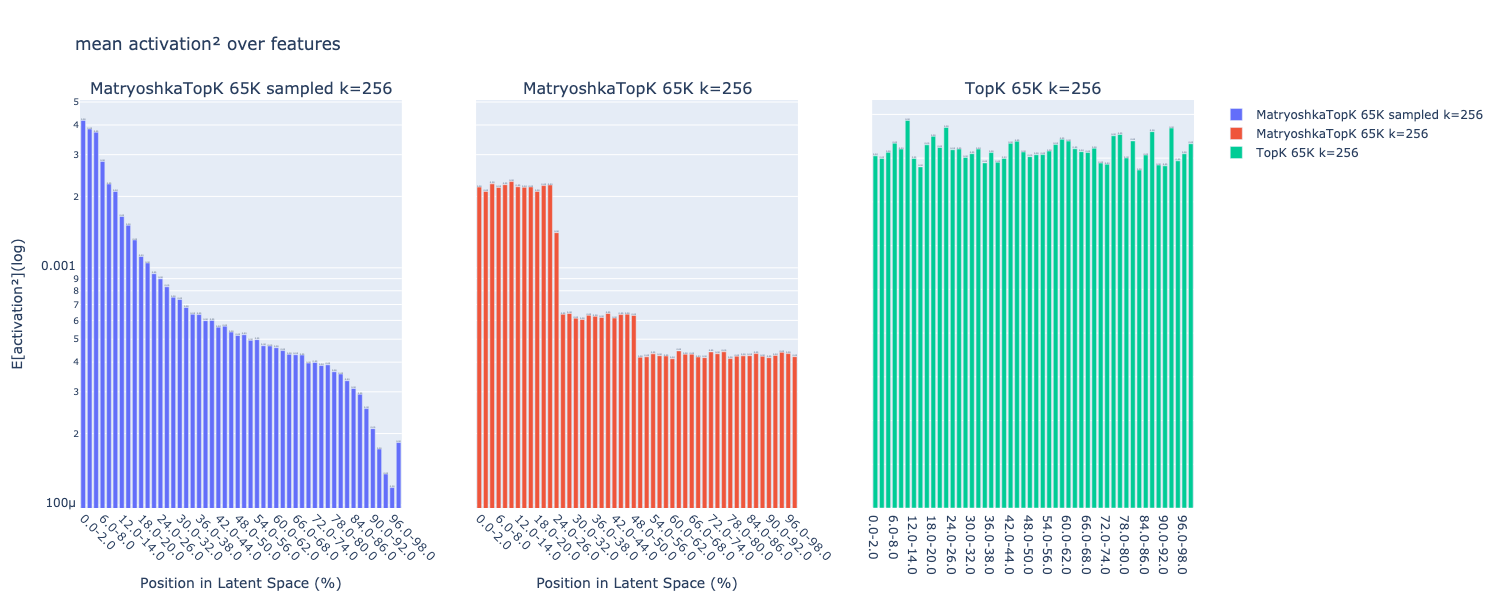}
    \caption{Mean activation squared by interval in latent space: sampled, non-sampled Matryoshka and baseline.}
    \label{fig:sampled_progressive_code_ce_loss}
\end{figure}

\bibliography{references}
\bibliographystyle{icml2025}

\newpage
\appendix
\onecolumn
\section{Feature Splitting}

Feature splitting is the phenomenon where as the dictionary grows in size, one basis vector gets decomposed into multiple separate basis vectors. 

In contrast, the standard TopK SAE exhibits a relatively uniform diagonal pattern, indicating that similar features tend to be distributed throughout the latent space with a natural locality which is likely a function of random initialization.

In contrast, the Matryoshka TopK SAE shows a distinctive stepped pattern, with clear discontinuities at the model's granularity boundaries $\{2^{14}, 2^{15}, 2^{16}\}$. This indicates that features within each granularity level form relatively isolated clusters, with limited similarity to features in other granularity levels. 

\begin{figure}[H]
    \centering
    \includegraphics[width=3in]{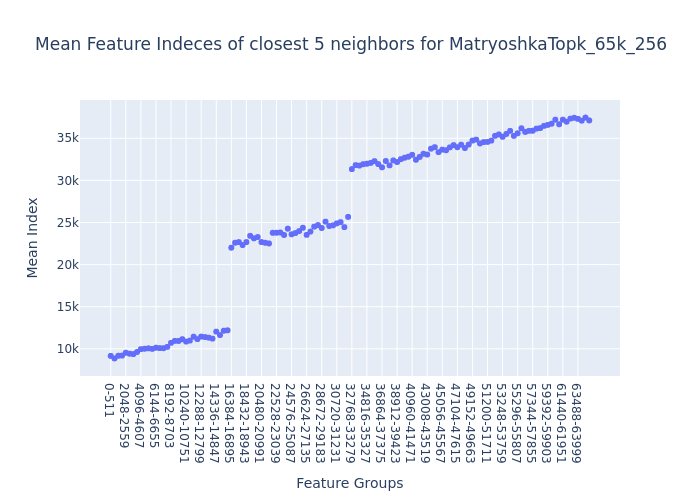}
    \caption{We compute the mean index of the top 5 closest feature for each feature for the Matryoshka TopK SAE}
    \label{fig:feature_splitting_Matryoshka_topk}
\end{figure}

\begin{figure}[H]
    \centering
    \includegraphics[width=3in]{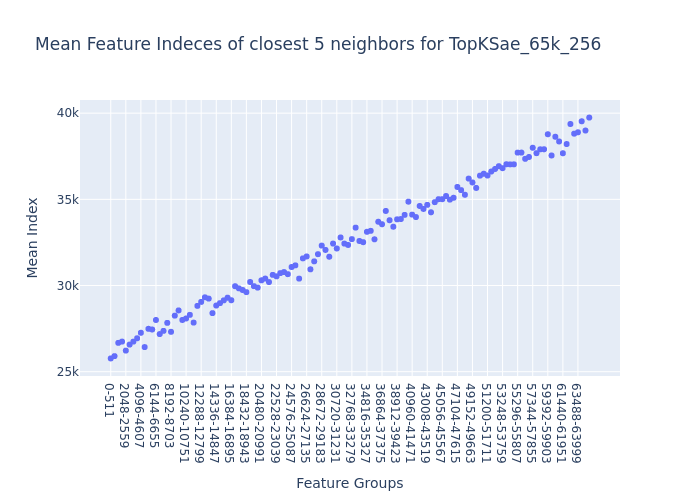}
    \caption{We compute the mean index of the top 5 closest feature for each feature for the TopK SAE}
    \label{fig:feature_splitting_topk}
\end{figure}

The natural locality of features in the standard TopK SAE can be attributed to the random initialization process, where nearby features in the latent space tend to develop related functionality during training.

\begin{figure}[H]
    \centering
    \includegraphics[width=3in]{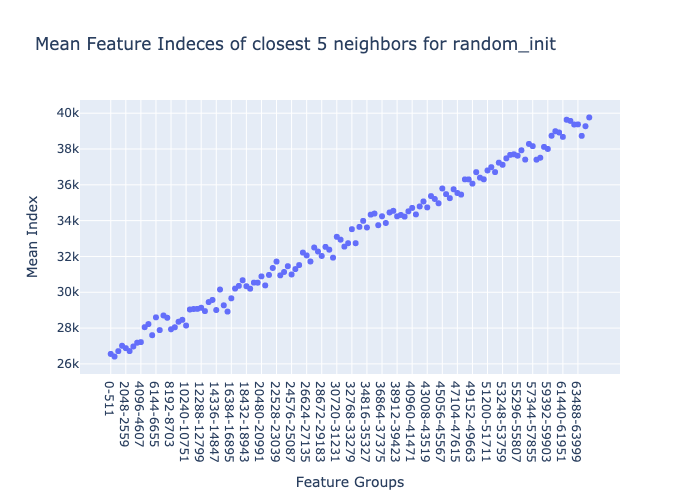}
    \caption{We compute the mean index of the top 5 closest feature for each feature for a random initialized decoder}
    \label{fig:feature_splitting_random_init}
\end{figure}

\section{Figures}

\begin{figure}[H]
    \centering
    \includegraphics[width=3in]{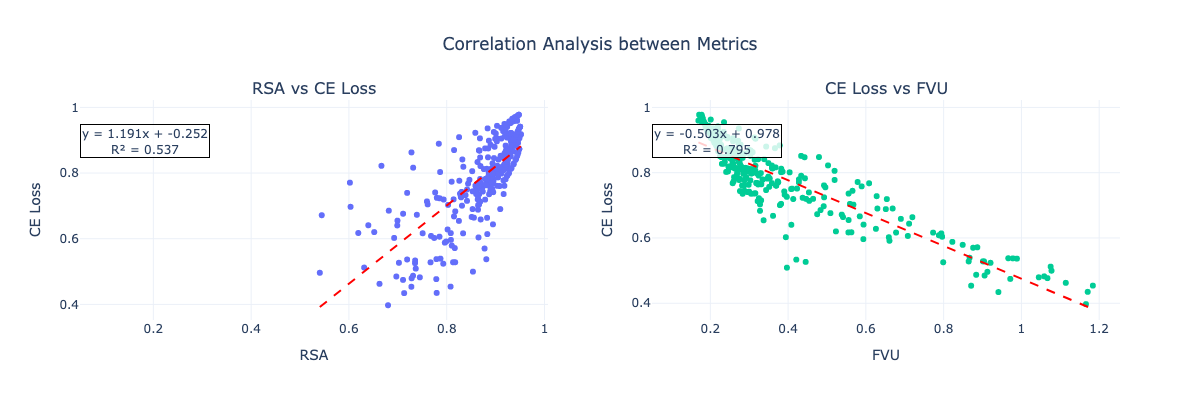}
    \caption{Correlation analysis between different evaluation metrics (FVU, CE Loss, and RSA). The scatter plots show pairwise relationships with linear regression fits, displaying both Pearson correlation coefficients (r) and coefficients of determination (R²).}
    \label{fig:metrics_correlation}
\end{figure}

\label{sec:limitations_and_future_work}
\section{Limitations and Future Work}

While our results are promising, it's important to note that our experiments were conducted on relatively modest-sized SAEs compared to recent work, scaling to tens of millions of features \citep{templeton2024scaling}\citep{gao2024scalingevaluatingsparseautoencoders}. Our methods remain to be validated at larger scales, though we find the observation that the dictionary power law holds at multiple scales\ref{dict_power_law} encouraging.

A key limitation in our implementation of Matryoshka SAEs lies in the decoder kernel \citep{gao2024scalingevaluatingsparseautoencoders}. However the kernel has not been optimized for performing multiple decoding passes per encoding 
step, leading to redundant computations as the decode kernel is invoked $|\mathcal{M}|$ times for $m \in \mathcal{M}$, separately computing $W_{\text{Dec}_{1:m}} topk(z_{1:m})$ for each granularity level. Given our weight sharing structure where $W_{\text{Dec}_{1:m_0}} \subseteq W_{\text{Dec}_{1:m_k}}$. Moreover it is highly likely that $topk(z_{1:m_0}) \subset topk(z_{1:m_1}) \dots \subset topk(z_{1:m_n})$. Thus a modified implementation could be meaningfully faster.

The challenge of feature-splitting presents another significant limitation. While permuting dictionaries by $E[activation^2]$ ordering provides a lightweight approach to distilling large pretrained SAEs, this method becomes less effective as the ratio of granularity to model size ($\frac{G}{N}$) decreases. This degradation occurs because a single feature in a small SAE is decomposed into multiple features in larger ones, and selecting only the most important of these split features fails to capture the complete functionality present in the original, unified feature. Future research could focus on developing efficient methods to recombine or "reverse" this feature-splitting during the distillation process, potentially through feature clustering or adaptive merging strategies.

To the best of our knowledge, we are the first to observe that as the decoding step in SAEs is highly sparse, for every sparse code, we can decode it multiple times using different parts of our dictionary with asymptotically negligible overhead. We consider other training approaches that apply these ideas highly promising and likely more computationally efficient than current methods.

\end{document}